\documentclass{article}
\usepackage{algorithm}
\usepackage{graphicx}
\usepackage{algpseudocode}
\usepackage[normalem]{ulem}
\usepackage{lipsum}
\usepackage{amsmath}
\usepackage{amssymb}
\usepackage{amsthm}
\newtheorem{definition}{Definition}[section]
\usepackage{tcolorbox}
\tcbuselibrary{listings, breakable}
\usepackage{wrapfig}
\usepackage[a4paper, margin=1in]{geometry}
\usepackage{mathpazo}
\usepackage{titlesec}
\usepackage{booktabs}
\usepackage{hyperref}
\titlespacing*{\section}{0pt}{10pt}{0.5em}
\titlespacing*{\subsection}{0pt}{8pt}{0.3em}
\usepackage{pifont}

\begin{document}

\title{{\huge\textsc{Lightning Grasp}} \\ \textsc{High Performance Procedural Grasp Synthesis \\ with Contact Fields}}

\author{Zhao-Heng Yin and Pieter Abbeel \\ \\  UC Berkeley EECS}
\date{}  % This removes the date from the title
\maketitle
\vspace*{-0.6cm}

\begin{abstract}
Despite years of research, real-time diverse grasp synthesis for dexterous hands remains an unsolved core challenge in robotics and computer graphics. We present Lightning Grasp, a novel high-performance procedural grasp synthesis algorithm that achieves orders-of-magnitude speedups over state-of-the-art approaches, while enabling unsupervised grasp generation for irregular, tool-like objects. The method avoids many limitations of prior approaches, such as the need for carefully tuned energy functions and sensitive initialization. This breakthrough is driven by a key insight: decoupling complex geometric computation from the search process via a simple, efficient data structure - the Contact Field. This abstraction collapses the problem complexity, enabling a procedural search at unprecedented speeds. We open-source our system to propel further innovation in robotic manipulation.
\end{abstract}

\begin{figure*}[h]
  \centering
  \includegraphics[width=\linewidth]{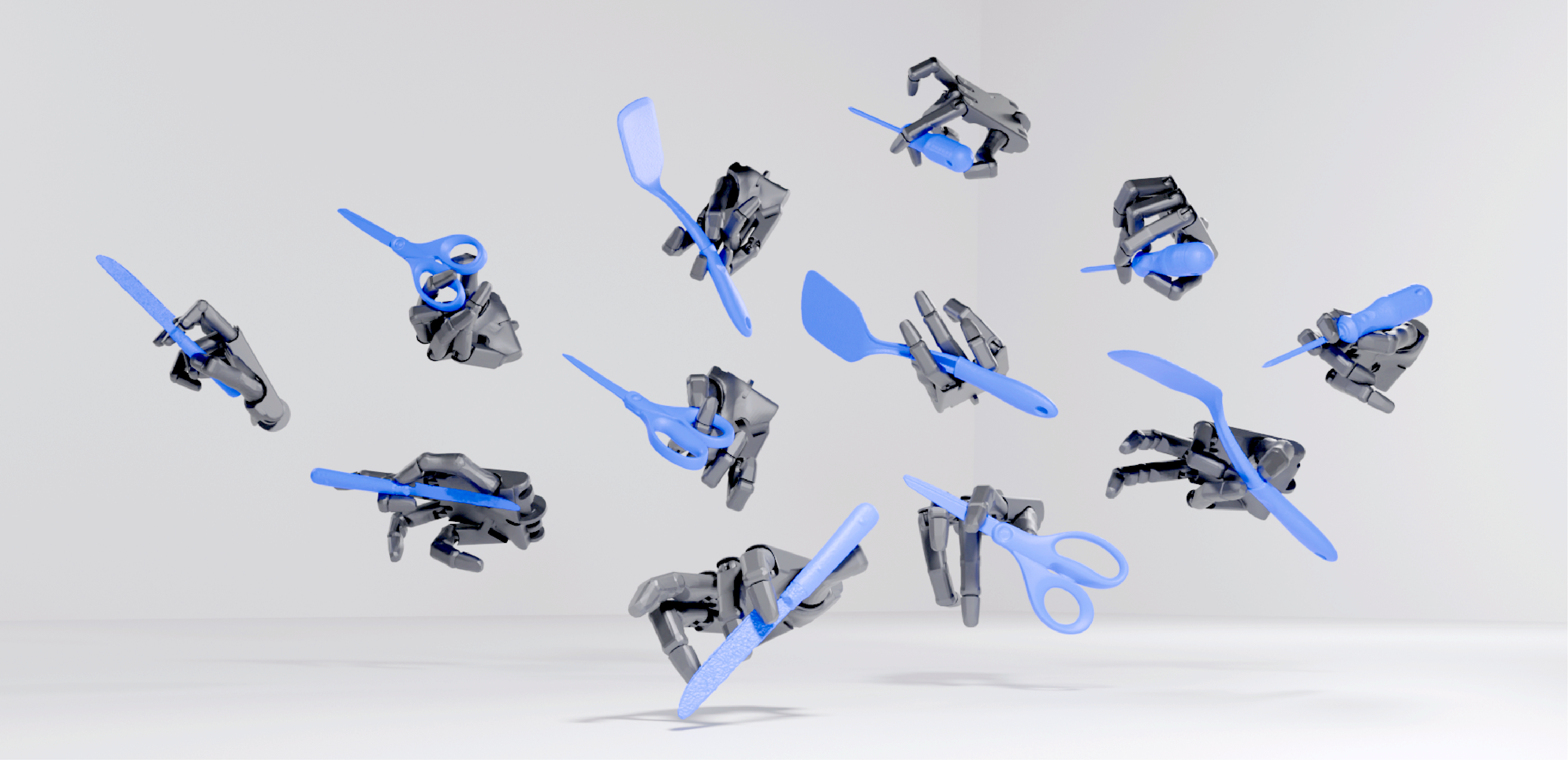}
  \setlength\tabcolsep{4.2pt}
   \begin{tabular*}{\linewidth}{@{\extracolsep{\fill}} l c c c c}
    \toprule
    Metric~(on 1 A100) & DexGraspNet~\cite{wang2022dexgraspnet} & SpringGrasp~\cite{chen2024springgrasp} & BODex~\cite{chen2025bodex} & Lightning Grasp~(Ours) \\
    \midrule
    Diverse Contact & \ding{51} & \ding{55}~(Fingertip) & \ding{55} ~(Fingertip) & \ding{51} \\
    Effective Sample/sec~($\uparrow$) & $<$3 & $<$3 & 30-50 & \textbf{300-1000} \\
    Forward Time~(sec)~($\downarrow$) & 1800-2000 & 10-40 & 100-120 & \textbf{2-5}\\
    \bottomrule
   \end{tabular*}

  \caption{Lightning Grasp is a high-performance procedural~(analytical) grasp synthesis algorithm. Compared to the other state-of-the-art analytical methods, it runs several orders of magnitude faster, produces a greater diversity of grasps, adapts to complex objects and high-DOF hands, and eliminates the need for the manual energy function tuning or template design. As shown in the figure, Lightning Grasp robustly handles highly irregular shapes with flexible, adaptable grasp poses within seconds. Code is available at \href{https://github.com/zhaohengyin/lightning-grasp}{\textcolor{blue}{https://github.com/zhaohengyin/lightning-grasp}}}.
  \vspace{-0.6cm}
  \label{fig:teaser}
\end{figure*}

\section{Introduction}
It is late 2025, three years after the surge of generative AI and two decades after GraspIt!~\cite{miller2004graspit}, yet a fast and effective procedural (analytical) grasp synthesis algorithm remains elusive. Procedural grasp synthesis algorithms serve as crucial data engines for developing data-driven grasping and manipulation policies~\cite{bohg2013data, khandate2023sampling, weng2024dexdiffuser, lin2024twisting, zhang2024dexgraspnet, yin2025dexteritygen}, in addition to their direct applications in robotics. However, as illustrated in Figure~\ref{fig:teaser}, recent methods~\cite{liu2021synthesizing, wang2022dexgraspnet, chen2024springgrasp, zhang2024dexgraspnet, chen2025bodex, lu2025grasping, chen2025dexonomy} remain either slow or limited in various aspects, and none are capable of real-time, diverse grasp synthesis. In Dexterity Gen~\cite{yin2025dexteritygen}, we developed a CPU-based~(mostly) grasp search algorithm for Anygrasp-to-anygrasp training. It ``worked'', but also needed a huge CPU cluster for parallel computing and many heuristics. To alleviate the challenges others might face in dexterous grasping and manipulation, and to question the dominance of parallel-jaw grippers, we set out to develop a more efficient method.

\begin{figure*}[t]
  \centering
  \includegraphics[width=0.9\linewidth]{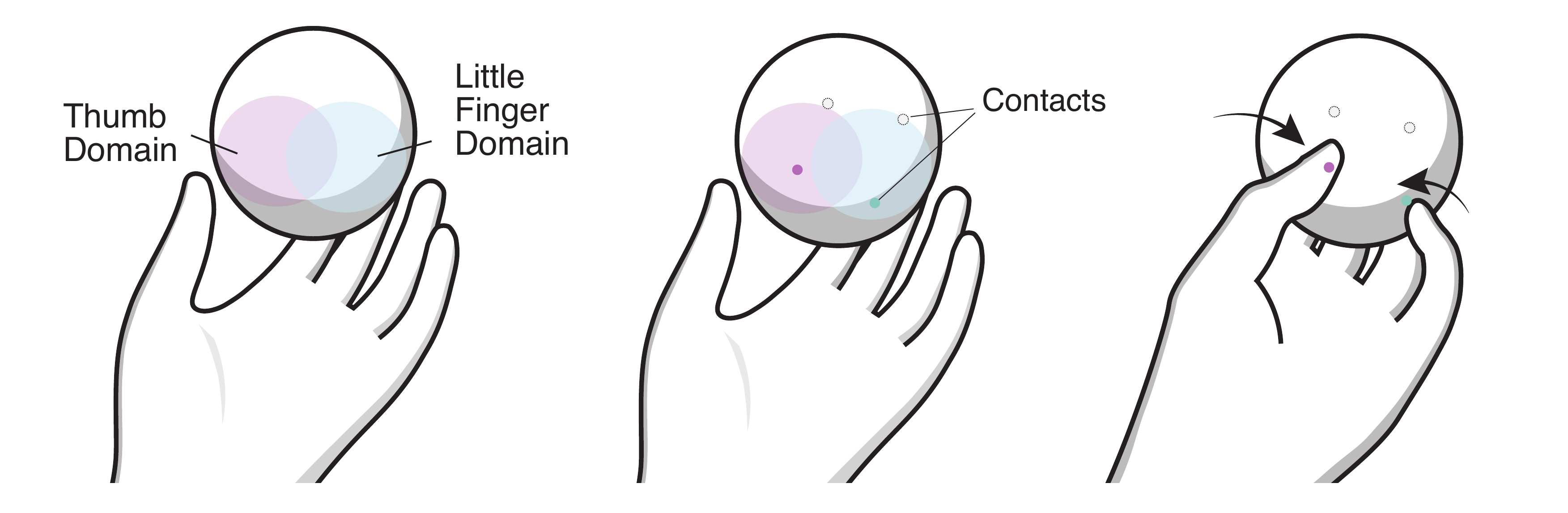}
  \caption{The Idea. Lightning Grasp follows three steps to grasp an object. (Left) We first identify the contact domains of each finger on the object surface, where each domain represents the feasible region a finger can reach. (Middle) We then search for an optimal set of contact points within these domains. (Right) Finally, the grasp is realized by positioning the fingers at the computed contact points.}
  \label{fig:idea}
\end{figure*}
We present Lightning Grasp, a novel algorithm for high-performance dexterous grasp synthesis for diverse hands and objects. On an A100 GPU, a single forward pass of Lightning Grasp generates between 1,000 and 10,000 diverse, valid grasps (depending on object complexity) within 2–5 seconds, achieving orders-of-magnitude speedups over prior methods. This efficiency also generalizes to legacy GPUs such as the TITAN X, where the method attains real-time inference performance. A performance-optimized mode, which reduces diversity and batch size, further cuts the forward pass by half. Beyond raw speed, Lightning Grasp eliminates key human bottlenecks: it requires no manually designed hand-initialization templates and is free from the sensitive objective-weight tuning endemic to existing methods.

The performance breakthrough of Lightning Grasp is rooted in a key observation: traditional grasp synthesis confounds two fundamentally different types of computing: geometric computation and search/optimization. This entanglement creates a major performance bottleneck, as the optimization procedure is constantly slowed by intensive geometric computations. To overcome this, we introduce a decoupled architecture. Our central innovation is the contact field, a simple but powerful data structure that can detect and represent all feasible contact regions on an object efficiently. By providing a clear interface between geometry and optimization, the contact field enables rapid, high-quality contact optimization and grasp synthesis without the overhead of complex geometric computations. The following sections will guide you through the technical details of this approach, showcase its performance on diverse objects and hand models, and explore the new research avenues it opens. 
\vspace{-0.3cm}
\section{Overview}
\vspace{-0.25cm}
How can we grasp an object above or around our palm? Just three steps (Figure~\ref{fig:idea}): (Left) identify feasible contact regions, (Middle) select points for a stable grasp, and (Right) execute the grasp with fingers.
\vspace{-0.1cm}

Lightning Grasp implements this simple three-stage approach. First, we introduce a contact field to reduce contact domain detection to a collision detection problem. Next, a block-wise zeroth-order optimization identifies stable contact points within these domains, which are finally achieved via iterative kinematic optimization. While conceptually straightforward, the method's performance hinges on critical low-level designs. We will first cover the necessary preliminaries (Section~\ref{sec:preliminary}), then delve into the contact field (Section~\ref{sec:cf}), and finally present the end-to-end pipeline (Section~\ref{sec:pipeline}).

\section{Preliminaries}
\label{sec:preliminary}
\subsection{Notations}
In this paper, a mesh $M$ is a 3-dimensional submanifold of $\mathbb{R}^3$. $\partial M$ is its boundary, a 2-dimensional manifold embedded in $\mathbb{R}^3$. ${\rm normal}(p, M)$ is the set of outer normal vectors of $M$ at $p$. 

\vspace{-0.2cm}
We define a hand (or any kinematic object) as $H=(H_M, \mathcal{C}, \mathcal{F})$. Here, $H_M \triangleq \{M_i\}_{i=1}^n$ is a collection of hand link meshes, $\mathcal{C}$ is the joint configuration space, and $\mathcal{F}$ denotes the forward kinematics function, which can produce the pose of any coordinate frame rigidly attached to $M_i$ under a given joint configuration $q \in \mathcal{C}$. 

\subsection{Grasp}
Given an object model mesh $O$ and a hand model $H$, a grasp is defined as a tuple $(P, q)$, where $P$ is the object's pose in the hand's frame, and $q\in \mathcal{C}$ is the hand's configuration~(i.e., joint positions). Furthermore, for a grasp to be valid, it should satisfy the following criteria. 
\vspace{-0.4cm}
\paragraph{1. No Penetrations} There should be no hand-object penetrations. Formally, let $H_M(q)$ be the hand mesh according to configuration $q$~(computed with forward kinematics $\rm FK$), we require that the contact set~(or manifold) $C(P,q)=H_M(q)\cap T(O;P)\subset \partial H_M(q)\cup \partial T(O;P)\subset \mathbb{R}^3$. Here, $T(O, P)=\{Px|x\in O\}$ denotes the object mesh under transform $P$. The condition states that two meshes intersect at their boundaries. In practice, however, we usually allow a small penetration margin around 2mm.
\vspace{-0.4cm}
\paragraph{2. Grasp Stability} The grasp should fulfill certain grasp stability conditions. There are many kinds of grasp stability metrics such as form and force closures. However, the closure condition is somewhat too strong and many human grasps~(e.g. two finger pinch grasp) do not satisfy these closure conditions either. Therefore, in this paper, we use the following self-balancing $\epsilon$-wrench setup from recent works~\cite{khandate2023sampling, chen2023synthesizing, yin2025dexteritygen}, but our approach extends to other metrics as well. Specifically, we require that there exists a contact point subset $C=\{p_1,p_2,...,p_k\}\subset C(P,q)$ and the associated normals $\{n_1, n_2, ..., n_k\}$ such that the solution to the following optimization problem is smaller than $\epsilon$:

\noindent\textbf{Frictionless Self-balancing Wrench Optimization~(FSWO)} $(\{p_i, n_i\}, \lambda)$%
\begin{equation}
\begin{aligned}
    &\underset{\alpha}{\text{minimize}} 
    && \left\Vert  \sum_{i=1}^n \alpha_i n_i \right\Vert^2 
    + \lambda \left\Vert  \sum_{i=1}^n \alpha_i (p_i \times n_i) \right\Vert^2 \\
    &\text{subject to} 
    &&  \exists j, \alpha_j = 1, \\
    &&& \alpha_i \geq 0,\quad \forall i = 1, \dots, n
\end{aligned}
\label{eqn:fswo}
\end{equation}

In the objective function above, the first term corresponds to the resultant force, and the second term corresponds to the resultant momentum. The condition states that there is a non-degenerating~(since $\alpha_j=1$) combination of finger forces $\alpha$ that can achieve a small enough resultant force and momentum. Note that the resultant momentum is dependent on the selection of the reference point in general. However, if the resultant force is equal to zero~(our case), it does not depend on the reference point. 

One may also be interested in considering friction with coefficient $\mu\geq 0$. We simply need to plug in friction terms into the objective:

\noindent\textbf{General Self-balancing Wrench Optimization~(GSWO)} $(\{p_i, n_i, x_i, y_i\}, \lambda, \mu)$%
\begin{equation}
\begin{aligned}
    &\underset{\alpha, \beta^{(x)}, \beta^{(y)}}{\text{minimize}} 
    && \left\Vert  \sum_{i=1}^n \alpha_i n_i + \beta^{(x)}_i x_i+\beta^{(y)}_i y_i\right\Vert^2 
    +\lambda \left\Vert  \sum_{i=1}^n p_i \times (\alpha_i  n_i + \beta^{(x)}_i x_i+\beta^{(y)}_i y_i) \right\Vert^2 \\
    &\text{subject to} 
    &&  \exists j, \alpha_j = 1, \\
    &&& \alpha_i \geq 0,\quad \forall i = 1, \dots, n \\
    &&& (\beta^{(x)}_i)^2 + (\beta^{(y)}_i)^2 \leq \mu^2 \alpha_i^2.
\end{aligned}
\end{equation}
$x_i$ and $y_i$ from an orthonormal basis of the tangent plane at $(p_i, n_i)$. Each $\beta$ corresponds to the strength of friction. Obviously, the optimum of GSWO is a lower bound of FSWO. Both problems can be decomposed into $n$ convex subproblems, which are efficiently solvable with projected gradient descent.

\subsection{Hardness of Grasp Synthesis}
Grasp synthesis is inherently challenging due to the high-dimensional search space. Nonetheless, it is useful to examine the key bottleneck in this problem. The main difficulty arises from the geometric constraint requiring exact contact between two complex meshes, compounded by the additional stability requirement for the contact points. Recent approaches typically model the no-penetration condition as a differentiable energy function $E_{pen}$ and use an attraction energy $E_{attract}$ to pull the hand toward the object surface. When carefully tuned, these terms create a delicate balance that positions the fingers precisely on the object. However, computing and optimizing these energies is computationally expensive due to mesh complexity, and the approach is highly sensitive to hyperparameters because the two energies counteract each other. As discussed earlier, the geometric constraint should be decoupled from the optimization process. To address this, we introduce the \textbf{contact field} in the next section.

\begin{figure}[t]
  \centering
  \includegraphics[width=\linewidth]{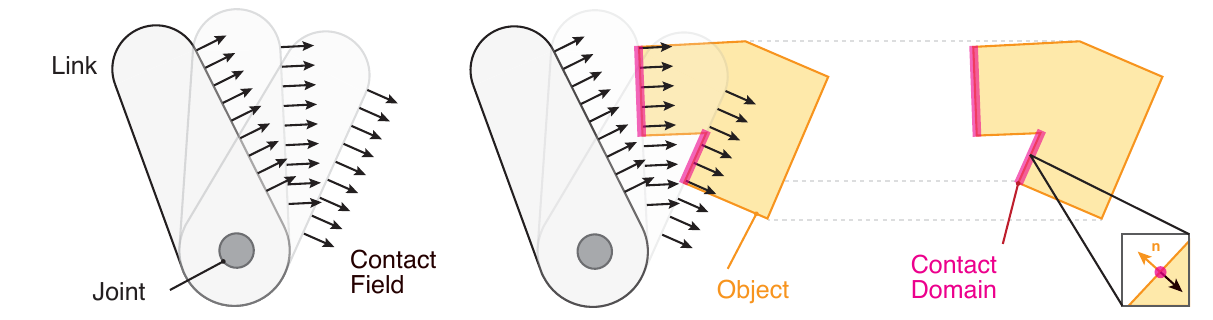}
  \caption{Contact Field and Its Interaction with Objects. A contact field is a collection of vectors in $\mathbb{R}^3\times \mathbb{S}^2$~(black arrows). This represents the potential contacts a hand can afford in space. Intersecting it with an object, we extract the contact domain on the object surface.}
  \label{fig:cf}
\end{figure}

\section{Contact Field}
\label{sec:cf}
\textit{Contact field} is the core data structure that allows us to simplify the grasp synthesis problem. We refer readers to Figure~\ref{fig:cf} for an intuitive illustration before the following formal definitions.

\subsection{Definitions}
A contact field characterizes spatial contacts that a hand can potentially generate. This is represented as a 6D geometry object, encoding both position and normal. We first define point-based contact field as follows.

\begin{definition}[Contact Field~(Point)] For a point $p\in\partial M_i$ with its associated normal $n\in {\rm normal} (p, M_i)$, its contact field in a given frame $B$ is defined as 
\begin{equation}
    CF_B(i, p,n)=\{ {\rm FK} ((p,n); i, q) |q \in \mathcal{C}\}\subset \mathbb{R}^3\times\mathbb{S}^2.
\end{equation}
Here, $\mathcal{C}$ is the joint configuration space. ${\rm FK}((p,n); i, q)$ computes the transformed $p$ and $n$ in frame B under joint configuration $q$, and this function is naturally induced by $\mathcal{F}$. Each ${\rm FK}((p,n); i, q)$ refers to a \textit{contact vector}, and $CF_B(i, p, n)$ is a collection of contact vectors. We also omit $B$ and use notation $CF(i, p,n)$ when $B$ is the base frame of hand $H$. 
\end{definition}
The contact field of a hand is a collection of point-based contact fields defined above.
\begin{definition}[Contact Field~(Hand)] Given a hand $H$, its contact field is defined as:
\begin{equation}
    CF(H)=\bigcup_{(i,p)\in\partial\hat H_{M}, n\in {\rm normal}(p, M_i)} CF(i, p,n) \subset \mathbb{R}^3\times\mathbb{S}^2,
\end{equation}
where $\partial\hat H_{M} \triangleq  \cup_{M_i\in H_M} \{(i, p) | p\in \partial M_i\}$ is the indexed collection of all the boundaries of link meshes.
\end{definition}

\begin{figure}[t]
  \centering
  \includegraphics[width=\linewidth]{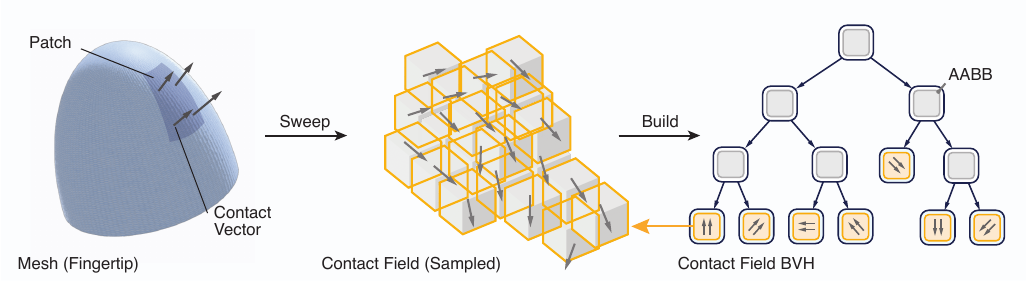}
  \caption{BVH Representation of Contact Field. (Left) We first decompose the hand link meshes into many small contact patches. (Middle) For each contact patch, we generate its contact fields by sampling joint configurations randomly and gathering corresponding contact vectors. (Right) We build a BVH to represent the sampled contact field. Each leaf node contains several normal vectors in the leaf box.}
  \label{fig:bvh}
\end{figure}

Finally, we introduce the following contact surface representation for an object mesh $M$.
\begin{definition}[Contact Surface Representation] Given a mesh $M$, its contact surface representation is defined as 
\begin{equation}
    S(M)=\{(p, -n)| p\in \partial M, n \in {\rm normal}(p, M)\}\subset \mathbb{R}^3\times\mathbb{S}^2.
\end{equation}
\end{definition}
\subsection{Contact Interaction}
With the definitions above, the potential contact interaction between an object mesh $O$ and the hand's contact field can be easily defined as $CF(H)\cap S(O)\subset \mathbb{R}^3\times \mathbb{S}^2$ which we call \textit{contact domain}. The contact domain encodes the tractable contact points on the object mesh surface. Extracting the first 3 cartesian dimensions will give us a tractable contact region on $\partial O$. Through the lens of the contact field, we can also understand why a soft, deformable human hand is so powerful: it can flexibly deform to provide diverse $\mathbb{S}^2$ configurations at each contact location, resulting in a much larger intersection $CF(H)$ and $CF(H) \cap S(O)$.

Clearly, this is a high-dimensional set intersection problem. The challenge is how to compute and use this interaction efficiently. We discuss an approximation to the contact field and contact domain in the next section.

\subsection{Implementation}
Similar to prior work on other 3D data structures, we first generate an approximation of $CF(H)$ using sampling. We randomly sample joint configuration $q\in\mathcal{C}$ and combine all the contact vectors under $q$. Then, to compute $CF(H)\cap S(O)$, a naive approach is to treat these spatial vectors as truncated rays and use them to hit the object surfaces and return the successful ones. However, this would result in too many unnecessary ray queries that do not fit into the GPU memory. This requires the opposite: using the object as query and organizing our contact field into an efficient data structure.

\paragraph{Contact Field BVH} We organize the sampled contact field with a Bounding Volume Hierarchy~(BVH). An illustration is shown in Figure~\ref{fig:bvh}. We split the space into several small boxes with each containing several contact normal vectors, and all the boxes are organized with a BVH. The construction process is summarized in Algorithm~\ref{alg:cfbvh}. 

\begin{algorithm}
\caption{BVH Construction of Contact Field}
\begin{algorithmic}[1]
\Require $n$ sampled contact vectors $X \subset  \mathbb{R}^3\times \mathbb{S}^2$. $w$ is box width.
\State Boxes $\{b_i=(l_i, h_i, S_i=[])\}\leftarrow$ GenerateBoxCover($X[:, :3]$, $w$);  \hfill // Grid cover.
\State $T\leftarrow$ LBVH($\{b_i\}$);   \hfill // Use LBVH~\cite{karras2012maximizing} construction.
\ForAll{$i \in \{1, \dots, {\rm len}(X)\}$} \textbf{in parallel}
    \State $I_i$ = BVHQuery($X_i.p, T$); \hfill // Return the indexes of all the hit boxes.
    \ForAll{$j \in I_i$}
        \State $S_j$.append($X_i.n$);    \hfill // Put contact vectors into corresponding boxes.
    \EndFor
\EndFor
\State (Optional) Build BVH for each $S_i$~(i.e. BLAS).
\State \Return{$T$}.
\end{algorithmic}
\label{alg:cfbvh}
\end{algorithm}
\paragraph{Object Contact Query} To compute an approximation of $CF(h)\cap S(O)$, we can randomly sample $(p, -n)$ from $S(O)$ and perform a collision check with the constructed BVH. During traversal in internal nodes, we use the cartesian position part $p$. As we reach a leaf node corresponding to box $b_i$, we check whether $-n$ aligns with one of the hand normal vectors in $S_i$ within some threshold~(i.e., $\exists x\in S_i, -x^Tn\geq \theta_{hit}$). The normal vector set $S_i$ on the leaf node can be organized with another BVH~(i.e. BLAS). However, as we find $|S_i|$ typically small (for instance, using 256 vectors to approximate $\mathbb{S}^2$ gives each vector a cone half-angle of about $7^\circ$), a simple brute-force dot-product check on GPU is already efficient.

\subsection{Fine-grained Contact Field}
We have illustrated an approach to compute the approximation of the entire contact field. However, this would result in a problem in later application: we do not know which point on which finger contributes to a specific feasible contact. This will be very inconvenient for kinematics optimization as we will need to assign a specific point on some finger to realize that contact. 

Therefore, we propose to use a fine-grained, decomposed contact field. We break down the whole hand surface into $m$ patches, and compute the contact field for each patch. This corresponds to break $\partial\hat H_{M}$ into several disjoint subsets $(\partial\hat H_{M})_i$. Then during query, we query these $m$ BVH separately and combine their result with tree index and box index. The decomposition is based on a simple, stochastic surface cover procedure. We keep choosing points on hand surface and use them to construct patch randomly until the whole surface is covered by patches. We understand that this covering procedure is suboptimal and does not yield the minimal number of patches. An optimal polynomial-time algorithm is left as an exercise for the reader.

\paragraph{Memory Consumption} A potential concern is about the spatial complexity of this approach. Here, we show its feasibility with an estimate. Assume that the box width is 1cm. Given that the moving range of each finger is typically bounded by a 30cm $\times$10cm $\times$ 10cm region, we need at most 3000 boxes for a contact field of 1cm$^3$ resolution. If a box holds 256 vectors, then all the vectors at the leaf node uses at most $3000\times 256\times16$B$\ \approx 12$MB data~(most of the boxes will contain no more than 50 vectors). For the bounding box and BVH metadata, we will need around 6000 nodes in total~(our BVH is a binary tree) and each node contains an AABB bounding box (32B) and the topology data (16B). These will consume $6000\times 48$B $\ \approx 0.3$MB. In total, we will need around 12MB at most for each BVH. Therefore, even 100 contact fields will consume at most $1.2$GB memory. The 3D normal vectors represented as float4 in leaf nodes can be further compressed to an 2 byte integer index~(8x memory reduction), since they share common values on a compact unit sphere. 

\begin{figure}[t]
  \centering
  \includegraphics[width=\linewidth]{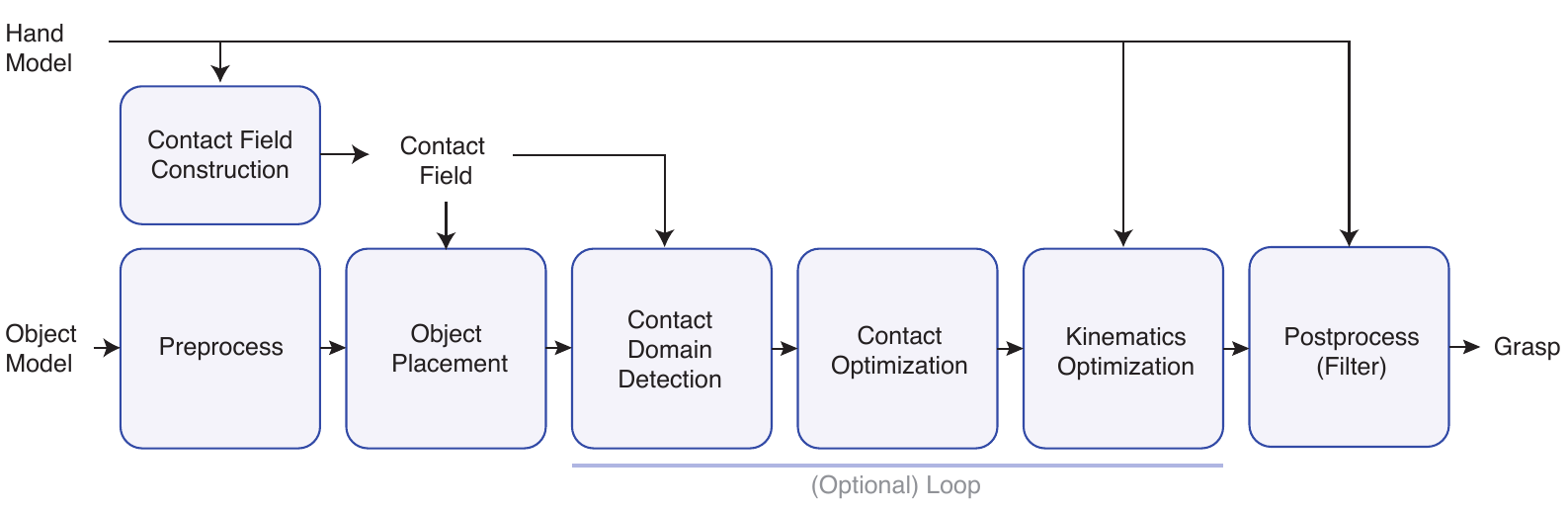}
  \caption{System Diagram. Our algorithm takes in hand model~(kinematics structure and mesh) and object model~(mesh) as input, and produces grasps. Our algorithm searches for object pose, contact point, and finger configurations sequentially. The output of each stage can be cached and reused for future forward search passes.}
  \label{fig:pipeline}
\end{figure}
\vspace{-0.3cm}
\section{Lightning Grasp Pipeline}
\label{sec:pipeline}
\vspace{-0.3cm}
Having established this powerful tool, we now present the full pipeline. We refer readers to Figure~\ref{fig:pipeline} for an overview. Our method takes in hand model~(mesh and URDF) and object model~(mesh) as input and outputs valid grasps. In this work, we implement all the components on NVIDIA GPUs, employing PyTorch~\cite{paszke2019pytorch} where possible~(e.g. kinematics) and writing custom CUDA kernels with C++ by ourselves when necessary~(e.g. collectives, BVH, and mesh-related operations). 

\subsection{Object Preprocessing}
We find it crucial to preprocess the object surface to reduce potential penetrations. For example, selecting contact points in highly-concave regions (e.g. unreachable holes) of the object surface significantly increases the risk of interpenetration. Therefore, we run a preprocessing procedure to remove the concave points from our object representation. We check whether placing a small box at an object point will lead to significant penetration. If so, we remove this object point from the candidates.

\subsection{Object Placement}
In the grasp search, we first generate the object pose. This decision is based on a simple observation: for most objects, as long as they are positioned somewhere above the palm, there is always a way to grasp them. In other words, our first expansion as object placement is very unlikely to fail. Conversely, searching for finger poses first makes it difficult to find a valid object pose, as the object often collides with the pre-placed fingers. We employ two main object placement strategies: an exhaustive, unsupervised method and an intuition-driven approach. These strategies yield distinct outcomes, making them suitable for complementary applications.
\vspace{-0.3cm}

\paragraph{Exhaustive Placement} We randomly choose a point in the contact field and align a randomly sampled object surface point with it. This strategy can produce very rare grasps, and we can use it to generate corner training/test cases. The trade-off is a potential loss in throughput, as some placements are too aggressive to be grasped reliably.

\paragraph{Canonical Placement} A more efficient approach is to specify a canonical object placement region, for instance, a box region above the palm specified. This approach trades the ability to generate rare grasp poses for significantly higher throughput with large aspect-ratio objects compared to the exhaustive method.

During object placement, we must also identify poses that establish contact with static links (e.g., the palm) to enable grasps like the power grasp. To do this, we initially place the object randomly against the static link surfaces with some probability. We then filter these placements, rejecting any that cause penetration and advancing only the successful ones—along with their corresponding contact vectors—to the subsequent finger pose generation stage.

\subsection{Contact Domain Generation}
\label{sec:pipeline3}
After we fix the object, we can employ the procedure in Section~\ref{sec:cf} to extract contact domains corresponding to each contact patch. To generate a grasp making $k$ object contacts, we collect $k$ contact domains from the results. A fundamental requirement is that these contact domains are independent, implying they originate from different fingers. This is because a single finger cannot kinematically realize two independent target contacts simultaneously in most cases. To generate $k$ independent contact domains, we first decide the dependency groups among the contact fields and merge the domains belonging to the same group. The dependency groups are determined by identifying the connected components of the kinematic tree after excluding all the static/fixed links. We randomly select $k$ contact domains from them and proceed to the contact point optimization. 

The inability to form multiple contacts on a single finger in a forward search step is an inherent constraint. To resolve this, an additional search phase is introduced to incorporate supplementary contact points. We have already applied this idea earlier: the static link contact vectors can be interpreted as the product of a prior contact search. In practice, however, a single forward search often proves sufficient to generate a diverse spectrum of viable grasps.

\begin{figure}[t]
  \centering
  \includegraphics[width=\linewidth]{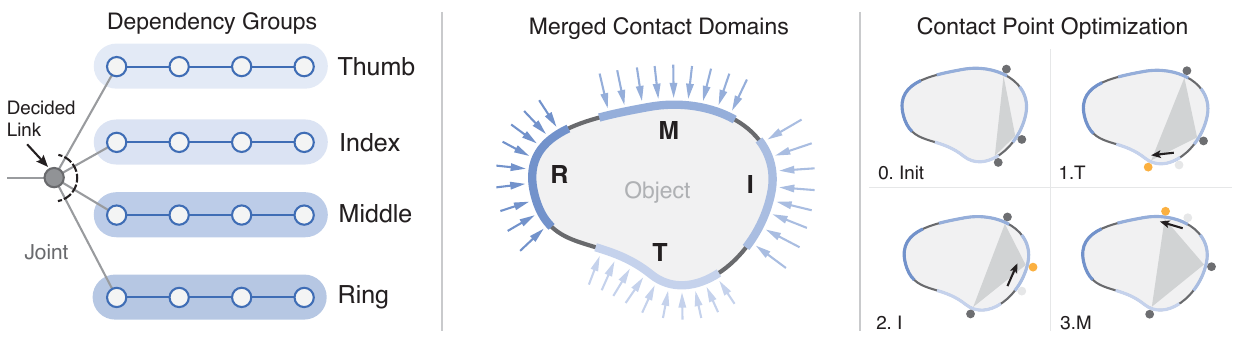}
  \caption{Contact Domain Generation and Contact Point Optimization. (Left) Dependency groups are connected components in the kinematics tree when we exclude the fixed/static links. (Middle) We compute a merged contact domain for each dependency group. (Right) We sample contact points in some of the merge contact domains~(Step 0) and perform iterative updates through zeroth-order optimization~(Step 1-3).}
  \label{fig:co}
\end{figure}
\subsection{Contact Point Optimization}
We search for one contact point in each of the contact domain to maximize the overall grasp quality objective. The optimization process can be formally written as 
\begin{equation}
\begin{aligned}
    &\underset{p_i,n_i}{\text{minimize}} 
    && J(p_1,n_1, ..., p_k, n_k) \\
    &\text{subject to} 
    && (p_i, n_i)\in \mathcal{D}_i.
\end{aligned}
\end{equation}
where $J$ is some grasp objective such as FSWO, and $\mathcal{D}_i$ is the contact domain. This leads to a bi-level optimization challenge as computing $J$ requires another low-level optimization. A simple method to optimize this non-differentiable objective is blockwise zeroth-order optimization. We optimize for one $(p_i, n_i)$ at one time through random local search. This turns out to be highly efficient as each $\mathcal{D}_i$ is essentially a 2D manifold~(Figure~\ref{fig:co}). Due to this, the overall process can quickly converge within 1 second. We show our algorithm in Algorithm~\ref{alg:opt}.
\begin{algorithm}[h]
\caption{Blockwise, Zeroth-Order Contact Point Optimization}
\begin{algorithmic}[1]
\Require Outer Iteration $n_o$, Inner Iteration $n_{in}$, Contact Domains $\mathcal{D}_i$~($i=1,2,...,k$).
\State $(p_i, n_i)\leftarrow$Random($\mathcal{D}_i)$.
\For{$it1\leftarrow1,2,...,n_o$}
    \For{$it2\leftarrow 1,2,..., k$}
        \State // Mutation Direction. $[..]$ is batched operation.
        \State $x, y\leftarrow {\rm Tangent}(n_i).$ \hfill //(returns an orthonormal basis of tangent plane).
        \State $[dx], [dy]\leftarrow$ Normal($n_{in}$, $\sigma^2$) $\times x$, Normal($n_{in}$,  $\sigma^2$) $\times y$.
        \State // Parallel Mutate
        \State $[p_i]'\leftarrow p_i+[dx]+[dy].$
        \State $[p_i', n_i'] \leftarrow$ Project$(p_i', \mathcal{D}_i)$.
        \State // Parallel Update
        \State $p_i, n_i \leftarrow \underset{(p, n)\in [p_i', n_i']}{\rm argmin} J(..., p_{i-1}, n_{i-1}, p,n, p_{i+1}, n_{i+1}, ...)$.
    \EndFor
\EndFor
\State \Return{$(p_1, n_1,...,p_k,n_k)$}.
\end{algorithmic}
\label{alg:opt}
\end{algorithm}

\paragraph{A Free Lunch} Our block-wise optimization procedure provides a computational 'free lunch' for grasp metrics based on optimization such as FSWO and GSWO. Recall that these optimization-based metrics take contact points as input. Since the contact points are changing slowly in high-level optimization, contact force solutions~(i.e. $\alpha$ in FSWO Objective~\ref{eqn:fswo}) obtained from previous low-level optimizers provide a high-quality initial configuration for the new low-level FSWO/GSWO problem. In this way, we can dramatically reduce the number of inner iterations to compute $J$.

\subsection{Kinematics Optimization}
\begin{figure}[t]
  \centering
  \includegraphics[width=\linewidth]{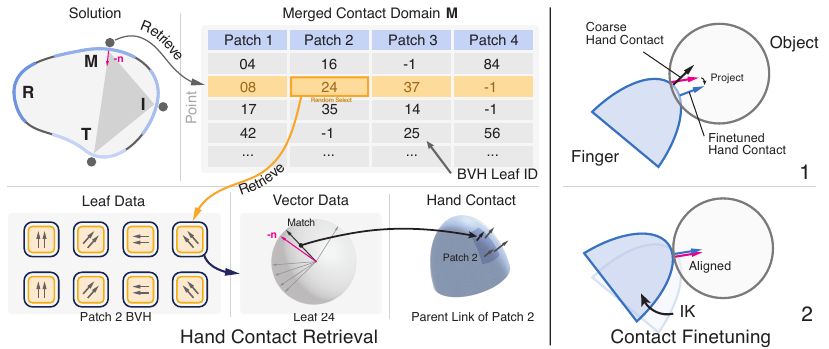}
  \caption{Operations in Kinematics Optimization. (Left) We use a reverse lookup through intermediate results to retrieve corresponding desired contact point on the hand mesh. Then we call IK to solve hand configuration. (Right) The original contact point may not realize the desired contact well. After initial IK, we finetune the hand contact point through projection during kinematics optimization.}
  \label{fig:ko}
\end{figure}

Having decided the object contact points, we can assign certain points on the finger surface to realize these contacts. As shown in Figure~\ref{fig:ko}, we first retrieve the corresponding contact point on the hand surface through a reverse lookup. We identify active patch-based contact fields at each object contact point, randomly pick one of them, and retrieve the closest aligned vectors inside the hit leaf node in the corresponding BVH.

Denote each retrieved contact point on hand as $(\tilde{p}_i, \tilde{n}_i)$. It has to realize $(p_i, n_i)$, or minimizing the distance between $p_i$ and $\tilde{p}_i$ and the angle between $n_i$ and $\tilde{n}_i$. The common 6D pose IK method does not apply to this problem, as the orientation update is not well-defined. We address this issue by framing the objective as two Cartesian position matching subproblems: we match $\tilde{p}_i$ to $p_i$ and $\tilde{p}_i + \beta\tilde{n}_i$ to $p_i+\beta n_i$, where $\beta$ is a small scalar to rescale the normal vector. At each step, we compute the joint update $\Delta q$ by optimizing the following damped least square~(DLS) problem:
\begin{align}
    \underset{\Delta q}{\text{minimize}}  \sum_{i}\left\Vert  \begin{bmatrix} \mathbf J_p(\tilde{p_i};q) \\ \mathbf J_p(\tilde{p}_i + \beta\tilde{n}_i;q) \end{bmatrix}  \Delta q - \begin{bmatrix} p_i - \tilde{p_i} \\ p_i+\beta n_i -  \tilde{p}_i + \beta\tilde{n}_i\end{bmatrix}\right\Vert^2 +\lambda\Vert \Delta q \Vert^2.
\end{align}
$\lambda$ is damping factor. $\mathbf J_p(\tilde{p_i};q)\in \mathbb{R}^{3\times {\dim \mathcal{C}}}$ denotes the position Jacobian for the contact point $\tilde{p}_i$, which can be computed given the common geometric Jacobian and the position of $\tilde{p}_i$ in its attached link frame. Let $\tilde{p}_i$ fixed to link $l_j$, which has a geometric Jacobian $\mathbf{J}(l_j; q) = \begin{bmatrix} \mathbf{J}_p \\ \mathbf{J}_r \end{bmatrix} \in \mathbb{R}^{6 \times \dim \mathcal{C}}$. Here, $\mathbf{J}_p$ and $\mathbf{J}_r$ are the linear and rotational components. Noticing the velocity relation $v_{\tilde{p_i}} = v_{l_j} + \omega_{l_j}\times (\tilde{p}_i)_{l_j}$ where $(\tilde{p}_i)_{l_j}$ is the position of $\tilde{p}_i$ in $l_j$ frame, we have
\begin{equation}
    \mathbf J_p(\tilde{p}_i;q)= \mathbf J_p -[(\tilde{p}_i)_{l_j}]_\times \mathbf J_r.
\end{equation}
$[]_\times$ is the matrix form of 3D cross product. We implement an inverse kinematics solver for multi-chain systems using PyTorch. Our solver also returns a binary mask to indicate unused joints. Their values will be decided later in another round of search.
\paragraph{Finetuning~(Phase II)} The procedure above~(Phase I) can work perfectly when we have an extremely fine-grained contact field representation. However, when using a low-resolution approximation~(i.e., large bounding boxes and contact patches), we require some additional finetuning. At each finetune step, we project each $p_i$ onto the latest target link to obtain an improved contact point $\tilde{p}_i$ on the target finger, and then call the DLS updates to finetune $q$. This alternating update process is repeated for several iterations.
\vspace{-0.3cm}

\subsection{Postprocessing}
In the final stage, we need to determine the joints whose values have not yet been fixed. For instance, if our grasp only involves thumb and index finger, we have to decide the position of middle finger and ring finger as well. In this open-sourced version, we assign random values to these joints and filter out those having hand-to-hand or hand-to-object collisions or not satisfying the grasp stability criterion. 

We implement two-phase collision detection with a AABB-based broad phase followed by different narrow phase detection algorithms based on geometry type. Specifically, we run convex decomposition over hand meshes and implement parallelized GJK~\cite{gilbert2002fast} algorithm for hand self-collision check. For point-based object representation, we use half-plane collision check to detect its penetration depth with respect to each hand link.

For collision-free grasps that lack stability, a more advanced approach is to perform an additional contact search using the unused fingers, as we already discussed in Section~\ref{sec:pipeline3}. This method enables the generation of multiple contact points on a single finger. This is the general form of Lightning Grasp, and will be integrated into a future software release.

\subsection{Discussion}
Viewing the algorithm through the lens of search reveals its broader utility and extensions. In this section, we review the core mechanics of Lightning Grasp from this perspective and discuss more advanced applications. Fundamentally, any search or problem-solving algorithm can be visualized as a tree. The process begins at a root node and expands by making decisions at each step; the sequence of decisions from root to leaf forms a complete solution to the problem: we decide object pose, contact fingers, contact points, and hand configuration sequentially at different tree layers. We implement the feasibility and stability constraints at the tree node expansion steps, which ensures the validity of our result. 

\paragraph{Completeness} 
\begin{wrapfigure}[12]{r}[0pt]{0.4\textwidth}
    \centering
    \includegraphics[width=\linewidth]{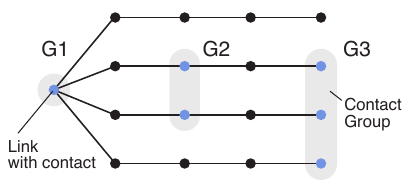} % replace with your image
    \caption{Any grasp can be decomposed into several contact point groups, whose points are independent of each other. Our algorithm (in general form) searches these contact groups to form a grasp.}
    \label{fig:side}
\end{wrapfigure}
A fundamental requirement for a search algorithm is completeness. Here, we define the completeness as the capability to find out all the potential solution. In this section, we mainly discuss the representation power of Lightning Grasp briefly, and show that any grasp lie on a potential search path of our algorithm. Given any stable grasp $g=(P, q)$, a sufficiently close approximation of $P$ is guaranteed with high probability through randomly sampling at our first step. Now we consider contacts and $q$. By definition, $g$ is supported by a collection of contact points attached to the hand kinematic tree for stability. We can partition the contact points into independent groups $G_1, G_2, ..., G_n$ such that:
\begin{enumerate}
    \item For any $e, e'\in G_i, e\neq e'$, after removing all links on the paths from the kinematics tree root to the parent links of \( G_1, G_2, \dots, G_{j-1} \),  $e$ and $e'$ belong to different connected components. 
    \item For any $e\in G_i, e'\in G_j$ with $i<j$, they must either reside on different subtrees, or $e'$ is a descendant of $e$.
\end{enumerate} 
This partition can be done trivially by a topological sort, and we omit its construction for brevity. The general form of our algorithm can produce $G_1$, $G_2$, ..., $G_n$ incrementally as discussed\footnote{However, to strictly guarantee this, contact point optimization should be applied only in the final round. This might be unnecessary in practice.}. Since $q$ is defined as an IK solution to $G_1, G_2, ..., G_n$ and our algorithm uses IK to realize these contacts, we can see that $q$ can be realizable by our approach when the initial guess of IK is sufficiently close to the desired $q$. 

\paragraph{Reusing Search Result} We can reuse the previous search results to speed up future grasp synthesis. For instance, we can resample contact points from a previously calculated contact domain, as one forward pass on a specific pose can still be far away from exhausting all the possible grasps. This approach is particularly suited for offline dataset generation, and we refer to it as multi-pass generation. From an algorithmic perspective, this is equivalent to another expansion and search starting from an internal node.

\paragraph{Data-driven Search} Although not implemented in this version, we believe that our search-based algorithm can be made even more efficient by incorporating self-play data as people did in model-based reinforcement learning. For example, we can train an object pose policy to generate candidate object poses, instead of relying on the human prior or fully random search. This can help to filter out bad object poses that are unlikely to have stable grasps given the current hand morphology. The training data can come self-play of Lightning Grasp, i.e., the model can learn which poses are promising given the past search experience. 

\begin{figure}[t]
  \centering
  \includegraphics[width=\linewidth]{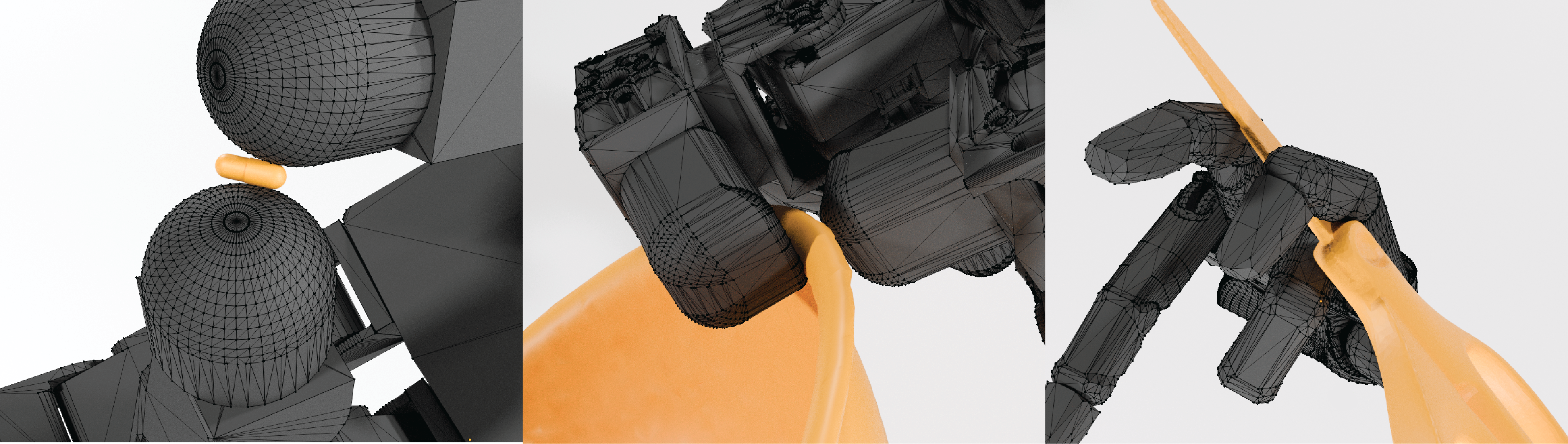}
  \caption{Making High-Quality Contacts. Our kinematics optimization procedure ensures precise contact between the fingers and the object's surface across diverse objects and hand morphologies.}
  \label{fig:synergy}
\end{figure}

\paragraph{Modularity} The modularity also enables interactive design, especially when we are interested in prompting some specific type of grasp. Specifically, we can manually designate object pose, contact patches, and allowed contact regions on the object surface before running a search. This can yield functional grasps matching human intention.

\begin{table}[htbp]
\centering
\caption{Amortized effective samples per second~(SPS) of our method on an NVIDIA A100 GPU. We evaluate representative everyday tools and objects spanning tiny~(e.g. capsule), regular~(e.g. apple) and non-convex~(e.g. cup and tools) categories. We show trimmed average $\mu$ (excluding minimum and maximum) in the last column. Baseline results are omitted since their SPS is at least 10× lower. All the configurations complete within 6 seconds.}
\vspace{0.15cm}
\label{tab:comparison}
\setlength\tabcolsep{4.2pt}
\begin{tabular}{lccccccccc}
\toprule
\textbf{Hand}   & \textbf{Capsule}  & \textbf{Apple} &\textbf{Spoon}  & \textbf{Cup}  & \textbf{Scissors} & \textbf{Screwdriver}&  \textbf{Plier}  & \textbf{Hammer} & \textbf{Trimmed $\mu$}\\
\midrule
Allegro & 1296.1 &  1578.8 &  955.6 &  1090.0 &  989.2 &  1020.6 &  1545.0 &  944.2 &  \textbf{1090.8} \\
LEAP & 3306.0 & 729.0 &  408.3 &  281.6 &  138.6 &  356.6 &  403.0 & 343.0 & \textbf{420.2} \\
Shadow  & 1060.2 & 288.4 &  329.4 &  181.5 &  416.2 &  895.0 &  745.1 & 678.6 & \textbf{558.8}\\
DClaw  & 2823.5 & 221.3 &  158.9 &  138.1 &  126.1 &  154.5 &  619.3 & 203.2 & \textbf{249.1}\\
\bottomrule
\end{tabular}
\label{tbl:result}
\end{table}

\begin{figure}[t]
  \centering
  \includegraphics[width=\linewidth]{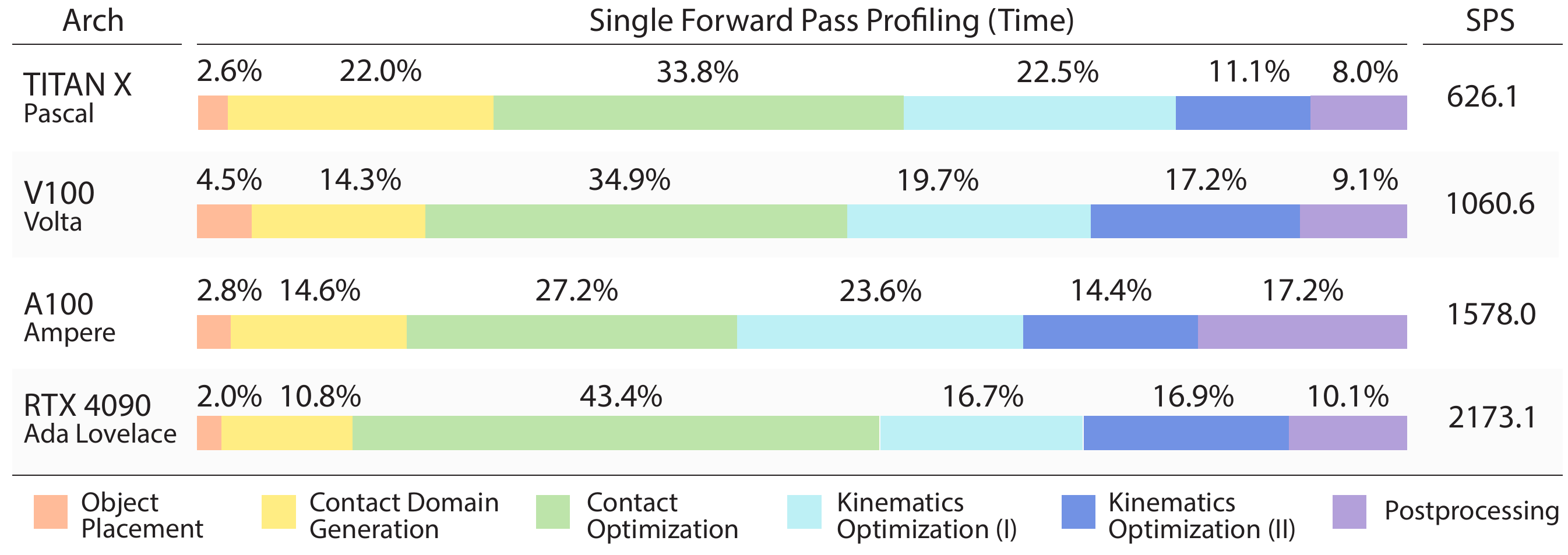}
  \caption{Profiling of a Single Forward Pass. Performance measured on an Allegro Hand grasping a YCB Apple. Workload is balanced across different GPU architectures, consistently achieving hundreds of samples per second (SPS). Notably, performance on a TITAN X GPU remains hundreds of times faster than an baseline running on an A100 GPU.}
  \label{fig:profile}
\end{figure}
\section{Results}
\subsection{Main Results}
\begin{figure*}[htbp]
  \centering
  \includegraphics[width=\linewidth]{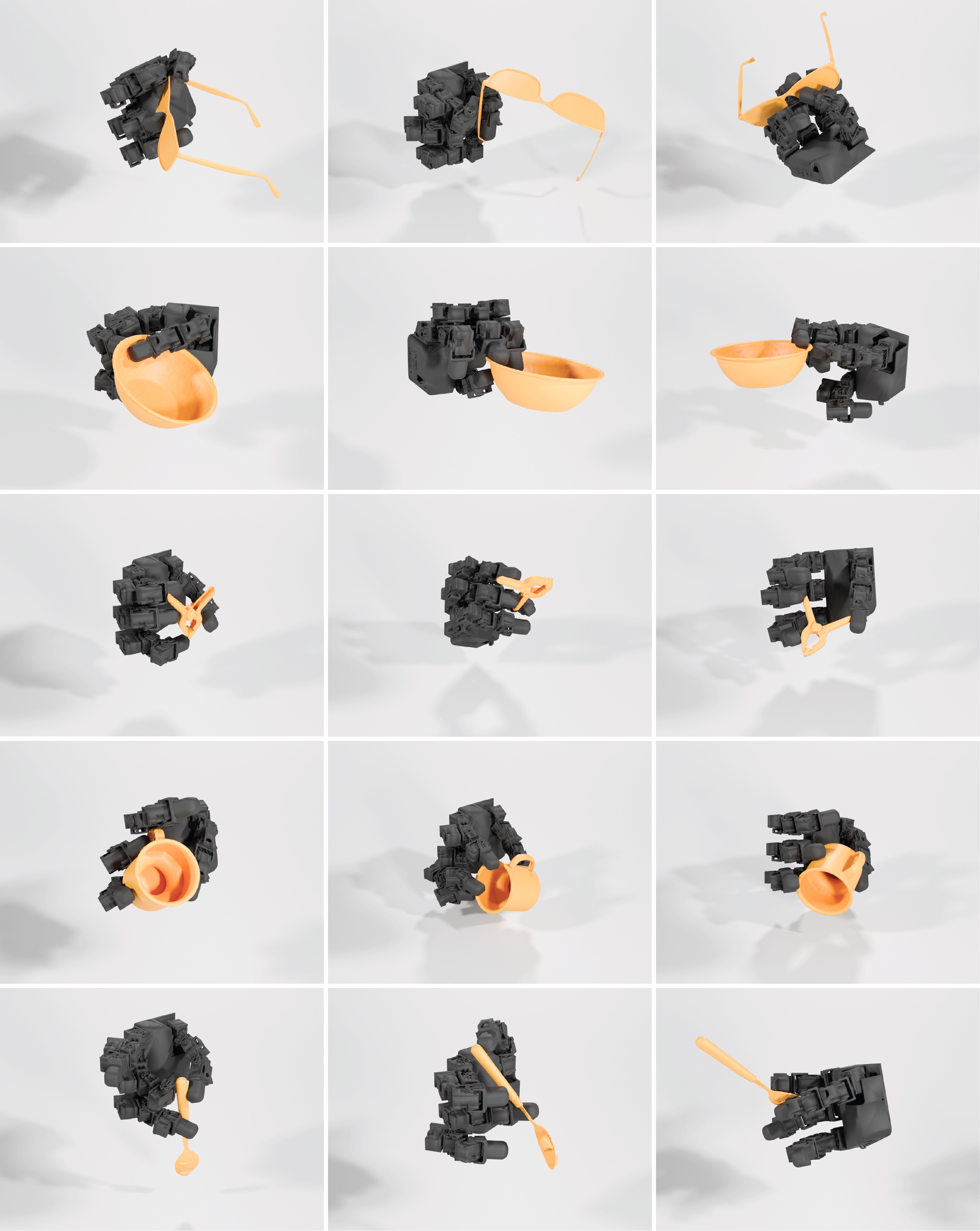}
  \caption{Random Grasp Synthesis Samples of the LEAP hand (16 DOFs, 4 Fingers). From Top to Bottom: Glasses, YCB Bowl, YCB Clamp, YCB Mug, and YCB Spoon.}
  \label{fig:leap}
\end{figure*}

\begin{figure*}[htbp]
  \centering
  \includegraphics[width=\linewidth]{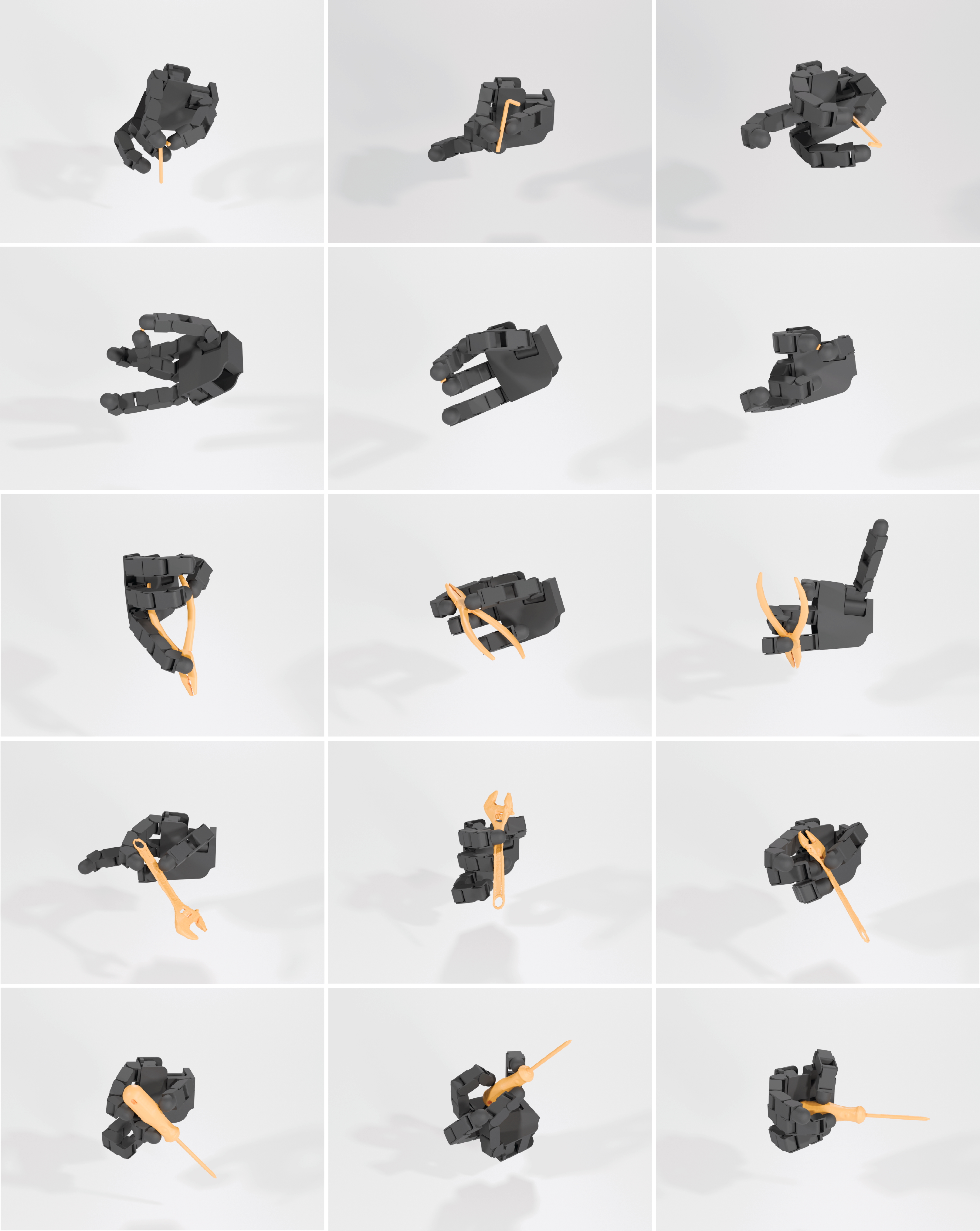}
  \caption{Random Grasp Synthesis Samples of the Allegro hand (16 DOFs, 4 Fingers). From Top to Bottom: Allen Wrench, Capsule, Plier, YCB Wrench, and YCB Screwdriver.}
  \label{fig:allegro}
\end{figure*}

\begin{figure*}[htbp]
  \centering
  \includegraphics[width=\linewidth]{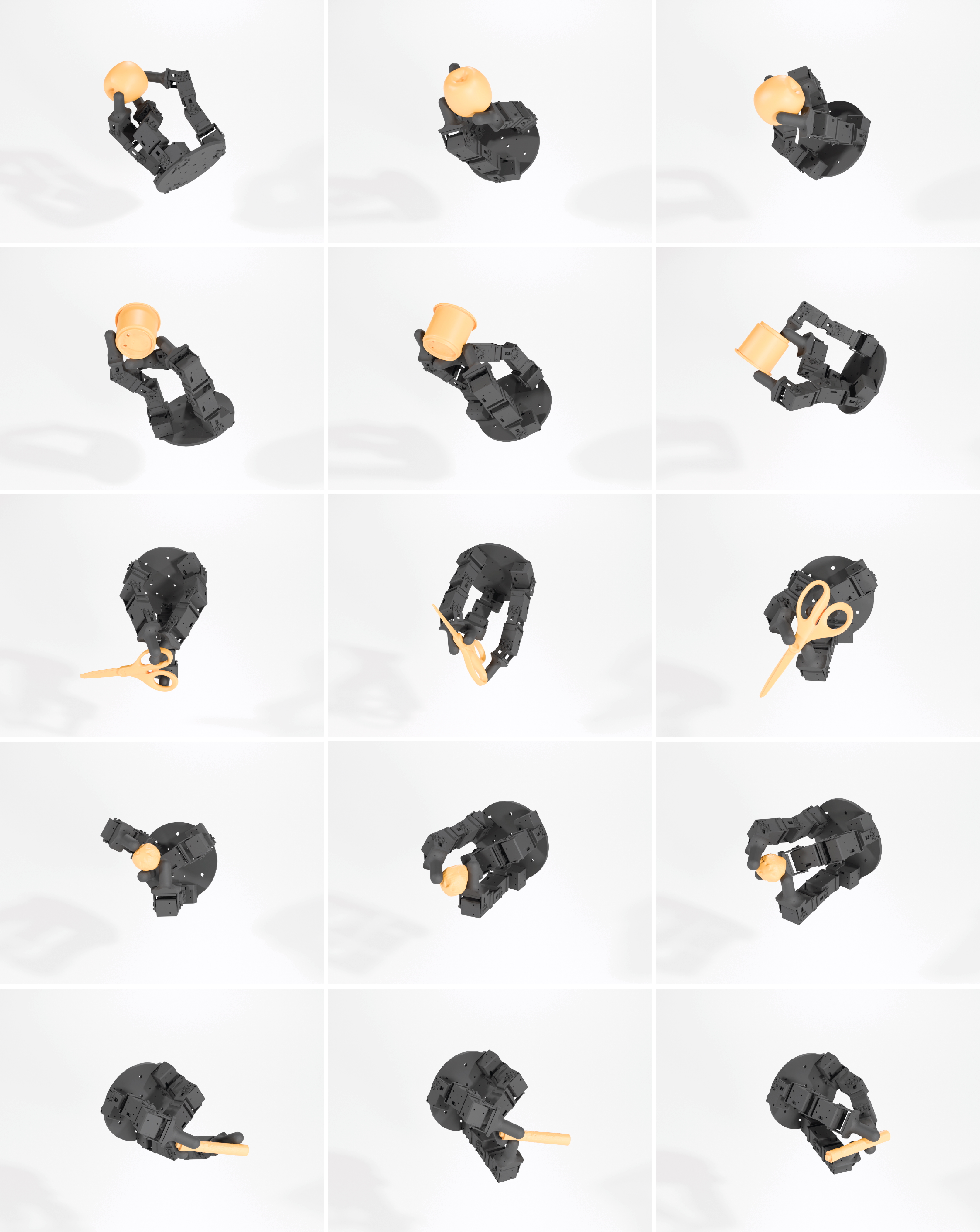}
  \caption{Random Grasp Synthesis Samples of the DClaw Gripper (9 DOFs, 3 Fingers). From Top to Bottom: YCB Apple, YCB Cup, YCB Scissors, YCB Strawberry, and YCB Marker. Its morphology differs significantly from that of a human hand. }
  \label{fig:dclaw}
\end{figure*}

The results are best understood visually, as shown in Figure~\ref{fig:teaser},\ref{fig:leap}, \ref{fig:allegro}, \ref{fig:dclaw}. We evaluate our system on the Shadow hand~\cite{shadowhand}~(22DOF) , LEAP hand~\cite{shaw2023leap}~(16DOF), Allegro Hand~\cite{allegrohand}~(16DOF), and DClaw Gripper~\cite{ahn2020robel}~(9DOF) with diverse YCB~\cite{calli2017yale} objects and other open-sourced 3D objects from the internet~(see acknowledgements). From tiny capsules and flat scissors to large non-covex bowls, our method generates diverse and secure grasps for a wide range of irregular objects and for many hands. In particular, we find our method able to produce its functional grasp poses for those large-aspect ratio, nonconvex functional tools as well, although we do not specify any prior knowledge. We also provide the average throughput of the commonly used dexterous research hand in Table~\ref{tbl:result}. Regardless of the object complexities, our algorithm consistently guarantees computational efficiency. Interestingly, we find that the search on the Allegro Hand consistently yields more valid samples than on other embodiments. In contrast, LEAP Hand, Shadow Hand, and DClaw Gripper exhibit more collisions during the filtering phase. The bulky motor layout of the LEAP Hand leads to frequent self-collisions. The high-DoF, five-finger design of the Shadow Hand introduces additional finger-crossing collision patterns. The nonconvex design of the DClaw's fingertips introduces excessive collisions, while its lower number of degrees of freedom (DoF) further restricts potential grasping solutions. These findings also suggest that our algorithm may also serve as a useful evaluator for hand hardware design. We leave a more detailed discussion on hard cases to the next section.
\begin{figure}[t]
  \centering
  \includegraphics[width=\linewidth]{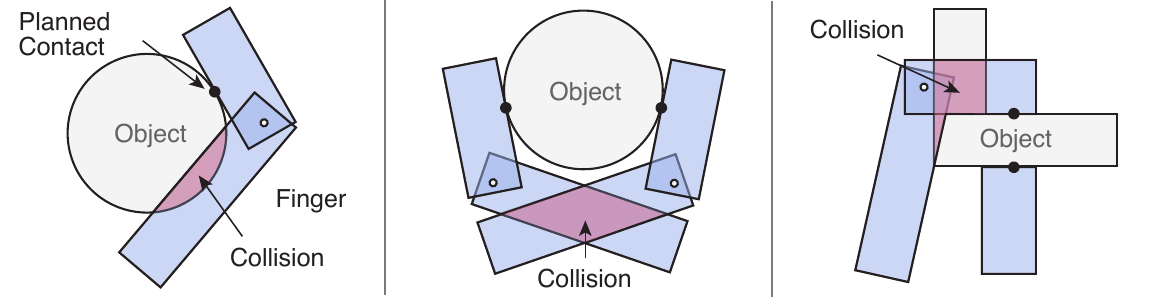}
  \caption{Common Failure~(Rejected) Samples Produced by Our Search. The cases shown on the Left and Middle are common across all test scenarios. However, the failure case on the Right, caused by the non-convex nature of the object, can significantly reduce the effective SPS. How to design data structures to prune these cases during search remains an open research problem.}
  \label{fig:failure}
\end{figure}
\subsection{Hard Case Analysis}
We observe that the effective SPS of our algorithm decreases significantly for certain types of objects, with the most challenging cases involving highly non-convex geometries, such as cups. A representative failure case is illustrated in Figure~\ref{fig:failure}~(Right). Although our kinematic optimization effectively resolves local collisions around each contact point (under the assumption of local convexity), global-scale penetrations can still occur. To eliminate this, one hypothesis is to add some form of finger shape information into each box of the contact field, and during contact query we can filter contact points leading to hand-object collisions. In short, the search in earlier stages should also be collision-aware. We leave this as a future work.

\subsection{System Performance Analysis}
Finally, we perform profiling to help readers identify potential computational bottlenecks in this system. We visualize our results on different main architectures, including Pascal, Volta, Ampere, and Ada Lovelace GPU architectures in Figure~\ref{fig:profile}. The workload is balanced in general across different components on various GPU architectures. Contact optimization and kinematics optimization each account for approximately 33\% of the total computation time. Our system's performance scales with modern hardware, achieving faster speeds on more advanced architectures. We also observe that our performance on TITAN X is already 20-100 times higher than that of existing methods on an A100 GPU.

\section{Conclusion}
In this paper, we presented Lightning Grasp, a conceptually simple yet high-performance algorithm for procedural grasp synthesis. We introduced a novel data structure, the contact field, which enables efficient decomposition and provides a practical solution to a challenge that has persisted for decades. We detailed the design of our system pipeline and its key components, highlighting both the conceptual framework and empirical performance of our approach. Our results demonstrated the speed, diversity, and robustness of the generated grasps. We believe that Lightning Grasp represents a significant step toward fully harnessing the potential of dexterous manipulators.
\section{Acknowledgments}
Zhao-Heng Yin is supported by the ONR MURI grant N00014-22-1-2773 at UC Berkeley. Pieter Abbeel holds concurrent appointments as a Professor at UC Berkeley and as an Amazon Scholar. This research was conducted at UC Berkeley and is not affiliated with Amazon.

The first author would like to thank Toru Lin for an interesting discussion on writing style. This paper is also dedicated to the Computer Science Divisions of the first author’s past institutions for the wonderful memories that came back during this project.

We would also like to thank the providers of the following 3D assets licensed under Creative Commons:
\begin{itemize}
    \item \texttt{https://sketchfab.com/3d-models/allen-wrench-02ebda14785f480b965da3fc2115cd7f}.
    \item \texttt{https://sketchfab.com/3d-models/pliers-63716d7381a742e3aef3d74c8759b0c9}.
    \item \texttt{https://sketchfab.com/3d-models/pill-capsule-93518f94ebd540c3b9da45ab67cf5590}.
    \item \texttt{https://sketchfab.com/3d-models/reading-glasses-cb92f0ac50ea46d5ab17036f279c3aa4}.
\end{itemize}
\bibliographystyle{plain}
\bibliography{reference}

@string{icra = "International Conference on Robotics and Automation~(ICRA)"}

@string{corl = "Conference on Robot Learning~(CoRL)"}

@string{irss = "{\rm In} Robotics: Science and Systems~(RSS)"}

@string{iicra = "{\rm In}  International Conference on Robotics and Automation~(ICRA)"}

@string{icorl = "{\rm In}  Conference on Robot Learning~(CoRL)"}

@string{inips = "{\rm In}  Neural Information Processing Systems~(NeurIPS)"}

@article{wang2022dexgraspnet,
  title={Dexgraspnet: A large-scale robotic dexterous grasp dataset for general objects based on simulation},
  author={Wang, Ruicheng and Zhang, Jialiang and Chen, Jiayi and Xu, Yinzhen and Li, Puhao and Liu, Tengyu and Wang, He},
  journal=iicra,
  year={2023}
}

@article{yin2025dexteritygen,
  title={Dexteritygen: Foundation controller for unprecedented dexterity},
  author={Yin, Zhao-Heng and Wang, Changhao and Pineda, Luis and Hogan, Francois and Bodduluri, Krishna and Sharma, Akash and Lancaster, Patrick and Prasad, Ishita and Kalakrishnan, Mrinal and Malik, Jitendra and others},
  journal=irss,
  year={2025}
}

@article{miller2004graspit,
  title={Graspit! a versatile simulator for robotic grasping},
  author={Miller, Andrew T and Allen, Peter K},
  journal={IEEE Robotics \& Automation Magazine},
  volume={11},
  number={4},
  pages={110--122},
  year={2004},
  publisher={IEEE}
}

@article{calli2017yale,
  title={Yale-CMU-Berkeley dataset for robotic manipulation research},
  author={Calli, Berk and Singh, Arjun and Bruce, James and Walsman, Aaron and Konolige, Kurt and Srinivasa, Siddhartha and Abbeel, Pieter and Dollar, Aaron M},
  journal={The International Journal of Robotics Research},
  volume={36},
  number={3},
  pages={261--268},
  year={2017},
  publisher={SAGE Publications Sage UK: London, England}
}

@article{liu2021synthesizing,
  title={Synthesizing diverse and physically stable grasps with arbitrary hand structures using differentiable force closure estimator},
  author={Liu, Tengyu and Liu, Zeyu and Jiao, Ziyuan and Zhu, Yixin and Zhu, Song-Chun},
  journal={IEEE Robotics and Automation Letters},
  volume={7},
  number={1},
  pages={470--477},
  year={2021},
  publisher={IEEE}
}

@article{paszke2019pytorch,
  title={Pytorch: An imperative style, high-performance deep learning library},
  author={Paszke, Adam and Gross, Sam and Massa, Francisco and Lerer, Adam and Bradbury, James and Chanan, Gregory and Killeen, Trevor and Lin, Zeming and Gimelshein, Natalia and Antiga, Luca and others},
  journal=inips,
  year={2019}
}

@article{gilbert2002fast,
  title={A fast procedure for computing the distance between complex objects in three-dimensional space},
  author={Gilbert, Elmer G and Johnson, Daniel W and Keerthi, S Sathiya},
  journal={IEEE Journal on Robotics and Automation},
  volume={4},
  number={2},
  pages={193--203},
  year={2002},
  publisher={IEEE}
}

@inproceedings{chen2025bodex,
  title={Bodex: Scalable and efficient robotic dexterous grasp synthesis using bilevel optimization},
  author={Chen, Jiayi and Ke, Yubin and Wang, He},
  booktitle=icra,
  year={2025}
}

@article{chen2024springgrasp,
  title={Springgrasp: Synthesizing compliant, dexterous grasps under shape uncertainty},
  author={Chen, Sirui and Bohg, Jeannette and Liu, C Karen},
  journal=irss,
  year={2024}
}

@article{chen2025dexonomy,
  title={Dexonomy: Synthesizing All Dexterous Grasp Types in a Grasp Taxonomy},
  author={Chen, Jiayi and Ke, Yubin and Peng, Lin and Wang, He},
  journal=irss,
  year={2025}
}

@inproceedings{chen2023synthesizing,
  title={Synthesizing dexterous nonprehensile pregrasp for ungraspable objects},
  author={Chen, Sirui and Wu, Albert and Liu, C Karen},
  booktitle={ACM SIGGRAPH 2023 Conference Proceedings},
  pages={1--10},
  year={2023}
}

@inproceedings{zhang2024dexgraspnet,
  title={Dexgraspnet 2.0: Learning generative dexterous grasping in large-scale synthetic cluttered scenes},
  author={Zhang, Jialiang and Liu, Haoran and Li, Danshi and Yu, XinQiang and Geng, Haoran and Ding, Yufei and Chen, Jiayi and Wang, He},
  booktitle=corl,
  year={2024}
}

@article{khandate2023sampling,
  title={Sampling-based exploration for reinforcement learning of dexterous manipulation},
  author={Khandate, Gagan and Shang, Siqi and Chang, Eric T and Saidi, Tristan Luca and Liu, Yang and Dennis, Seth Matthew and Adams, Johnson and Ciocarlie, Matei},
  journal=irss,
  year={2023}
}

@article{lu2025grasping,
  title={Grasping a Handful: Sequential Multi-Object Dexterous Grasp Generation},
  author={Lu, Haofei and Dong, Yifei and Weng, Zehang and Pokorny, Florian and Lundell, Jens and Kragic, Danica},
  journal={IEEE Robotics and Automation Letters},
  year={2025},
  publisher={IEEE}
}

@article{lin2024twisting,
  title={Twisting lids off with two hands},
  author={Lin, Toru and Yin, Zhao-Heng and Qi, Haozhi and Abbeel, Pieter and Malik, Jitendra},
  journal=icorl,
  year={2024}
}

@article{bohg2013data,
  title={Data-driven grasp synthesis—a survey},
  author={Bohg, Jeannette and Morales, Antonio and Asfour, Tamim and Kragic, Danica},
  journal={IEEE Transactions on robotics},
  volume={30},
  number={2},
  pages={289--309},
  year={2013},
  publisher={IEEE}
}

@article{weng2024dexdiffuser,
  title={Dexdiffuser: Generating dexterous grasps with diffusion models},
  author={Weng, Zehang and Lu, Haofei and Kragic, Danica and Lundell, Jens},
  journal={IEEE Robotics and Automation Letters},
  year={2024},
  publisher={IEEE}
}

@inproceedings{ahn2020robel,
  title={Robel: Robotics benchmarks for learning with low-cost robots},
  author={Ahn, Michael and Zhu, Henry and Hartikainen, Kristian and Ponte, Hugo and Gupta, Abhishek and Levine, Sergey and Kumar, Vikash},
  booktitle=corl,
  year={2020}
}

@article{shaw2023leap,
  title={Leap hand: Low-cost, efficient, and anthropomorphic hand for robot learning},
  author={Shaw, Kenneth and Agarwal, Ananye and Pathak, Deepak},
  journal={arXiv preprint arXiv:2309.06440},
  year={2023}
}

@misc{shadowhand,
  author       = {Shadow Robot Company},
  title        = {Dexterous Hand Series},
  url          = {https://shadowrobot.com/dexterous-hand-series/}
}

@misc{allegrohand,
  author       = {Wonik Robotics Co. Ltd.},
  title        = {Allegro Hand},
  url          = {https://www.allegrohand.com/}
}

@inproceedings{karras2012maximizing,
  title={Maximizing parallelism in the construction of BVHs, octrees, and k-d trees},
  author={Karras, Tero},
  booktitle={Proceedings of the Fourth ACM SIGGRAPH/Eurographics Conference on High-Performance Graphics},
  pages={33--37},
  year={2012}
}

\end{document}